\documentclass{article}
\usepackage{ijcai24}

\usepackage{times}
\usepackage{soul}
\usepackage{url}
\usepackage[hidelinks]{hyperref}
\usepackage[utf8]{inputenc}
\usepackage[small]{caption}
\usepackage{graphicx}
\usepackage{amsmath}
\usepackage{amsthm}
\usepackage{booktabs}
\usepackage{algorithm}
\usepackage{algorithmic}
\usepackage[switch]{lineno}
\usepackage{indentfirst}
\usepackage{multirow}
\usepackage{array} 

\title{Prompt-enhanced Network for Hateful Meme Classification}

\author{
Junxi Liu$^1$\and
Yanyan Feng$^1$\and
Jiehai Chen$^1$\and
Yun Xue$^{1}$\thanks{Corresponding author}\And
Fenghuan Li$^{2*}$\\
\affiliations
$^1$School of Electronics and Information Engineering, South China Normal University, Foshan 528225, China, \\
$^2$School of Computer Science and Technology, Guangdong University of Technology, Guangzhou 510006, China\\
\emails
\{liujunxi,fengyanyan,cjh\_scnu,xueyun\}@m.scnu.edu.cn,
fhli20180910@gdut.edu.cn
}

\begin{document}

\maketitle

\begin{abstract}
    The dynamic expansion of social media has led to an inundation of hateful memes on media platforms, accentuating the growing need for efficient identification and removal. Acknowledging the constraints of conventional multimodal hateful meme classification, which heavily depends on external knowledge and poses the risk of including irrelevant or redundant content, we developed Pen—a prompt-enhanced network framework based on the prompt learning approach. Specifically, after constructing the sequence through the prompt method and encoding it with a language model, we performed region information global  extraction on the encoded sequence for multi-view perception. By capturing global information about inference instances and demonstrations, Pen facilitates category selection by fully leveraging sequence information. This approach significantly improves model classification accuracy. Additionally, to bolster the model's reasoning capabilities in the feature space, we introduced prompt-aware contrastive learning into the framework to improve the quality of sample feature distributions. Through extensive ablation experiments on two public datasets, we evaluate the effectiveness of the Pen framework, concurrently comparing it with state-of-the-art model baselines. Our research findings highlight that Pen surpasses manual prompt methods, showcasing superior generalization and classification accuracy in hateful meme classification tasks. Our code is available at https://github.com/juszzi/Pen.
\end{abstract}

\begin{figure}
    \centering
    \includegraphics[width=1\linewidth]{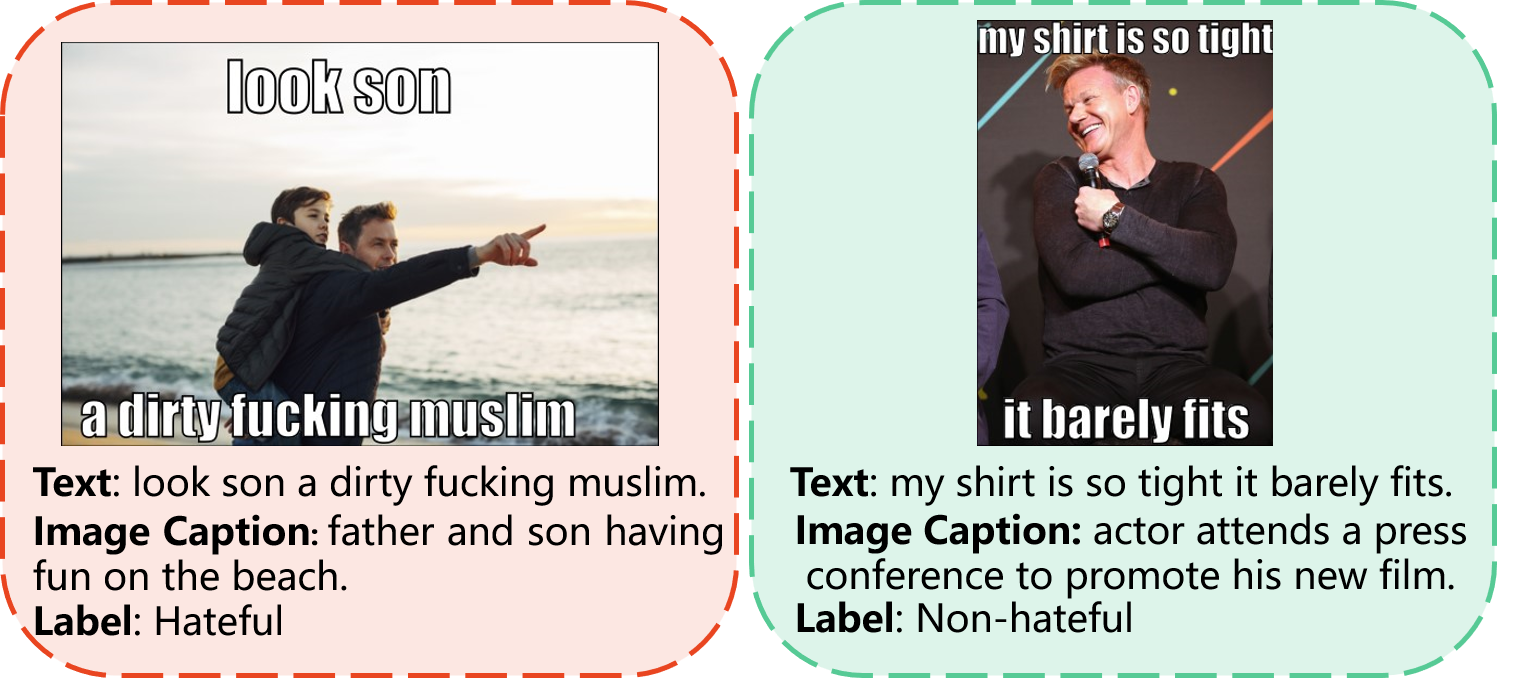}
    \caption{The red box indicates a sample labeled as ``hateful'', while the green box indicates a sample labeled as ``non-hateful''.}
    \label{fig:enter-label}
\end{figure}

\section{Introduction}

\noindent With the evolution of the internet, social media has emerged as the primary mode of communication, information sharing, and expressing opinions.  The rise of social media has introduced a new multimodal entity – memes, comprised of images and short texts. While this form has gained popularity on social media networks, it has also become a tool for some users to disseminate hate speech, causing serious harm to vulnerable groups \cite{DBLP:journals/chb/PiccoliCGSB24}. Due to the rapid dissemination of hateful memes, there is an urgent need to develop accurate classification methods. Figure 1 illustrates examples of hateful and non-hateful memes.

To address this issue, early efforts emphasized the alignment and fusion across modalities to recognize hateful memes in social media
\cite{DBLP:journals/corr/abs-2012-08290,DBLP:journals/corr/abs-2012-07788,DBLP:journals/corr/abs-2012-12975}. 
 Recognizing the need for intricate reasoning and contextual background knowledge in determining hatred in memes, attempts were made to enhance model classification accuracy by integrating external tools \cite{DBLP:conf/icmcs/ZhouCY21} or incorporating additional external knowledge \cite{DBLP:conf/mm/LeeCFJC21} within the visual language model framework.
Building upon this foundation, subsequent research began considering a modality transformation perspective. \citeauthor{DBLP:conf/emnlp/CaoLC022} transformed meme images into image captions, employing prompting methods and introducing external knowledge to guide Pre-trained language models (PLM) in predictions \cite{DBLP:conf/emnlp/CaoLC022}. 
In the latest work, building upon their prior work \cite{DBLP:conf/emnlp/CaoLC022}, \citeauthor{DBLP:conf/mm/CaoHKCL023} enhanced the quality of image Captions. They employed zero-shot visual question answering (VQA) with pre-trained vision-language models (PVLMs) for generating image captions \cite{DBLP:conf/mm/CaoHKCL023}. This enhancement led to superior image caption quality, achieving a state-of-the-art results in the current domain.

However, recent strategies for classifying hateful memes tend to emphasize improving model performance through the incorporation of additional external knowledge, potentially neglecting issues related to irrelevant or redundant content within such knowledge\cite{DBLP:conf/mm/LeeCFJC21,DBLP:conf/emnlp/BlaierMW21,DBLP:conf/emnlp/CaoLC022,DBLP:conf/uic/FangZHH22}. For instance, incorporating image entity recognition information \cite{DBLP:conf/mm/LeeCFJC21} may introduce entities that are unrelated or redundant to hateful memes, thereby adding irrelevant details that could interfere with the model's classification judgment. While some studies utilize prompting methods to guide PLM in leveraging external knowledge \cite{DBLP:conf/emnlp/CaoLC022}, this approach predominantly focuses on the data processing stage. It enhances the contextual learning capabilities of language models for classification by introducing prompt template tokens and demonstrations of different categories to the original sequence. However, it does not comprehensively address the training conditions of the sequence in the feature space.

Hence, our focus lies in extracting valuable information through a simple and effective network mechanism, enabling the PLM to adaptively select pertinent information for hateful meme classification. 
Existing prompt method guides PLM in classification by providing demonstrations corresponding to each label. Given the demonstrations for each label, there should be specific feature-level connections in the feature space between the contextual information of inference instances and the contextual information corresponding to the demonstrations of their correct labels.
Building upon this concept, we extend prompt method into the feature space, introducing a novel framework called the \textbf{P}rompt-\textbf{en}hanced network for hateful meme classification (\textbf{Pen}). In this framework, we initially process the sequences of the input PLM with prompts, followed by region segmentation. We extract global information features from both inference instance and demonstration regions, incorporating the prompt-enhanced multi-view perception module. This module perceives the global information features of inference instances and demonstrations from multiple views to make hate emotion judgments, enhancing the model's classification accuracy by effectively utilizing contextual information in input sequences. To better capture the relationships between hate and non-hate in the feature space, we introduce contrastive learning and adapt it to our framework, forming prompt-aware contrastive learning. This adaptation enhances the quality of the feature distribution for samples. In summary, the primary contributions of this paper are as follows:
\begin{itemize}
    \item We propose a model framework named Pen, which extends the prompt method into the feature space. By incorporating multi-view perception of inference instances and demonstrations in the feature space, Pen enhances hate classification accuracy, thereby improving the utilization of sequences.
    \item We propose a contrasting learning method compatible with manual prompting to align and differentiate sample features used for hate judgment. This method sharpens the features of samples from different categories, thereby improving the accuracy of the classifier.
    \item Through extensive ablation experiments conducted on two publicly available datasets, we validate the effectiveness of the prompt-enhanced framework, demonstrating its superiority over state-of-the-art baselines.
\end{itemize}

\section{Related Work}

\subsection{Multimodal Hateful Meme Classification}
\noindent Multimodal hateful meme classification aims to detect hateful implications in both text and images within memes. This task was initially introduced by the Hateful Memes Challenge (HMC) competition \cite{DBLP:conf/nips/KielaFMGSRT20}. Researchers, including \citeauthor{DBLP:conf/nips/KielaFMGSRT20}, conducted a series of experiments on this dataset, involving both unimodal and multimodal models, with superior performance observed in multimodal approaches. Subsequent studies \cite{DBLP:conf/acl-trac/SuryawanshiCAB20,DBLP:journals/corr/abs-2012-07788,DBLP:journals/corr/abs-2012-08290,DBLP:journals/corr/abs-2012-12975,DBLP:conf/emnlp/PramanickSDAN021} delved into exploring enhanced modality fusion methods, incorporating features learned from text and visual encoders using attention mechanisms and other fusion techniques.

Given the complexity of inferring hate in memes and the need for contextual background knowledge, recent research has started exploring approaches that integrate external knowledge to assist in hateful meme classification. Attempts have been made to augment the model's input with relevant external knowledge \cite{DBLP:conf/icmcs/ZhouCY21,DBLP:conf/mm/LeeCFJC21} to enhance the classification and interpretability of hateful content. \citeauthor{DBLP:conf/emnlp/CaoLC022} transformed meme images into image captions, introducing external knowledge and using prompting methods to guide PLM in prediction \cite{DBLP:conf/emnlp/CaoLC022}.
In the latest approach, \citeauthor{DBLP:conf/mm/CaoHKCL023} improved the quality of image Captions based on modality transformation, employing zero-shot VQA with PVLMs for image. This enhancement achieved SOTA results in the current research landscape. However, recent solutions, while supplementing external knowledge for assisting model judgments, still overlook irrelevant or redundant content within this knowledge. Thus, the effective and adaptive utilization of such external knowledge remains an urgent issue.

\subsection{Prompt For Hateful Meme Classification}

\noindent The natural approach to creating manual prompts involves using prompts that include task-specific descriptions and textual demonstration in a natural language manner as inputs for the model. For instance, in the case of sentiment classification for the movie review ``The film offers an intriguing what-if premise'', a prompt template, such as ``the sentiment of this review is [mask]'', can be added during data processing. Positive and negative examples are then appended to the sequence after the augmented prompt, allowing the language model to classify the [mask] token. A successful classification result should indicate a positive sentiment. \citeauthor{DBLP:conf/emnlp/CaoLC022} pioneered the use of prompt methods in multimodal hateful meme classification to guide PLM in hate reasoning. In a study by \citeauthor{DBLP:journals/corr/abs-2308-05596}, the combination of large language models and prompt learning was explored to address toxic content detection, demonstrating the effectiveness of prompt methods in hate detection tasks \cite{DBLP:journals/corr/abs-2308-05596}. Despite manual prompts being proven to solve various tasks with considerable accuracy \cite{DBLP:journals/csur/LiuYFJHN23}, manual prompt methods only process the input sequence to guide the natural language reasoning ability of PLM. The uncertainty remains about whether models can effectively assimilate sequences enhanced through prompt methods. Therefore, enhancing the model's utilization of sequence information remains a pressing issue.

\subsection{Contrastive Learning}
\noindent In the field of natural language processing, contrastive learning has gained significant traction in various studies\cite{DBLP:conf/acl/ZhangZZ0L22,DBLP:conf/naacl/JianGV22a,DBLP:conf/acl/LiangZL000X22,DBLP:conf/sp/QuHPBZZ23}. 
\citeauthor{DBLP:conf/acl/ZhangZZ0L22} introduced contrastive learning in multitask pretraining, leveraging unlabeled data clustering to obtain self-supervised signals and achieving optimal results in intent detection tasks. \citeauthor{DBLP:conf/naacl/JianGV22a} combined contrastive loss from prompt-based few-shot learners with standard Masked Language Modeling (MLM) loss. \citeauthor{DBLP:conf/acl/LiangZL000X22} applied target-aware prototype-based contrastive learning in zero-shot stance detection tasks.Recently, 
\citeauthor{DBLP:conf/sp/QuHPBZZ23}
explored the application of contrastive learning for hate classification, utilizing a multimodal contrastive learning model to unsupervisedly identify the primary groups associated with potential hateful memes.
Currently, contrastive learning has demonstrated significant effectiveness across various tasks, improving model performance in classification tasks. However, the current approaches primarily rely on simple label-oriented sample feature clustering or sample-driven self-supervised contrastive learning. The goals of contrastive learning applications are relatively narrow. Exploring additional forms of contrastive learning methods to enhance sample features is expected to enable models to learn diverse information, thereby improving model performance.

\section{Methodology}
\noindent \textbf{Problem Definition}
We define the hateful meme classification task as a series of binary tuples represented by the entity $M=\{T, I\}$, where text $T$ and image $I$ are interrelated. Our objective is to train a model to assess these tuples and output either ``hateful'' or ``non-hateful''. Following the framework proposed by \citeauthor{DBLP:conf/emnlp/CaoLC022}, we transform the multimodal setting into an unimodal one. Utilizing the image-to-text tool ClipCap\cite{DBLP:journals/corr/abs-2111-09734}, we convert images into image captions. After concatenating the text with the image caption, we introduce a prompt template ``it was [mask]'', along with demonstrations and external knowledge, incorporating them into a PLM. The language model then evaluates the [mask] token, selecting the appropriate label as the output.

In this study, our core idea is to extend the prompt method, applying the concept of prompt methods in the feature space to strengthen the connection between inference instances and demonstrations. By incorporating information from the entire sequence, we aim to improve the classification effectiveness of the language model.
In this section, we will provide a detailed overview of our approach to handling the hateful meme classification task. Figure 2 illustrates the structure of our proposed Pen framework, comprising Regional Information Global Extraction (Section 3.1), Prompt-enhanced Multi-view Perception (Section 3.2), and Prompt-aware Contrastive Learning (Section 3.3).
\begin{figure*}
    \centering
    \includegraphics[width=0.96\linewidth]{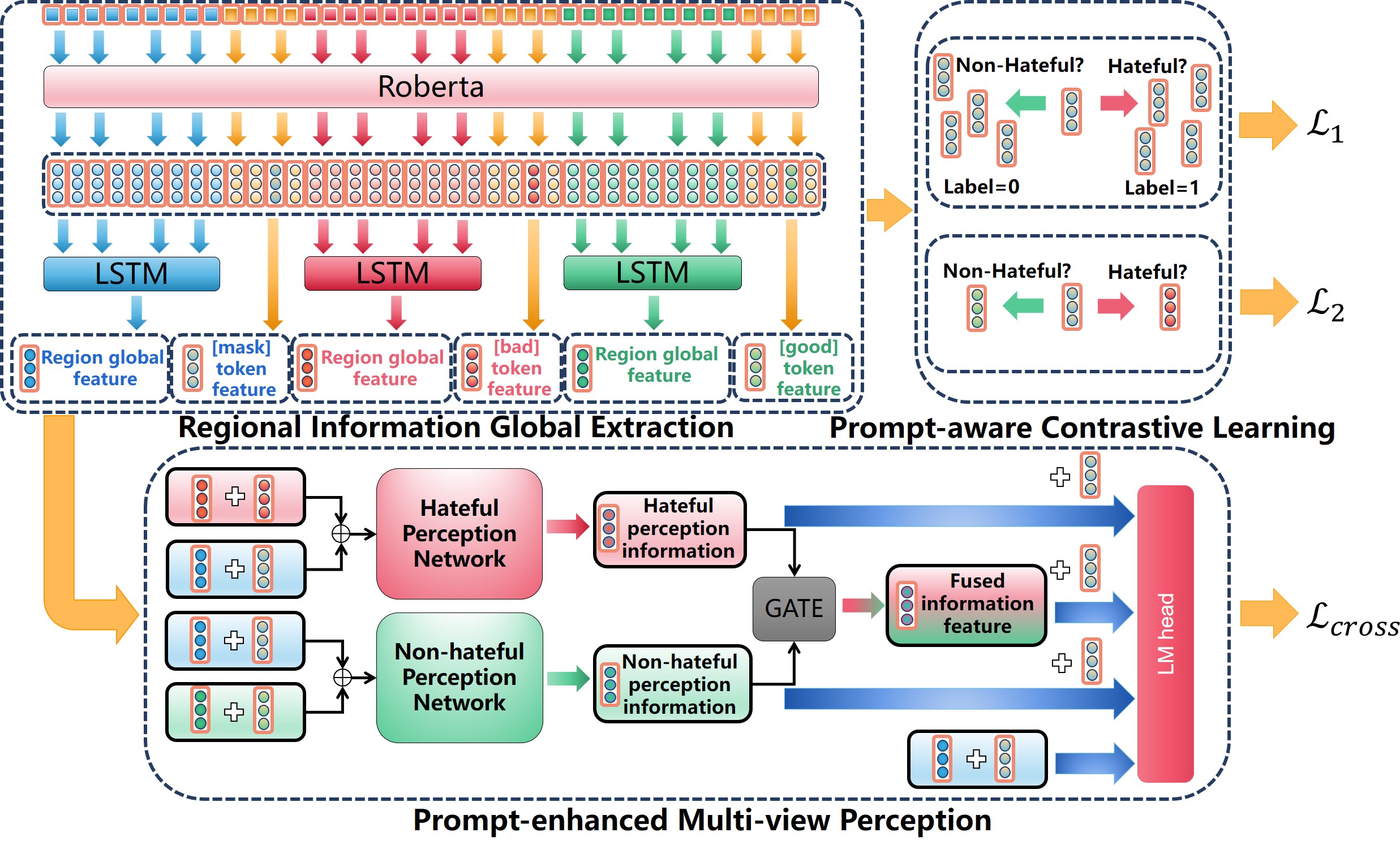}
    \caption{Overview of Pen Framework}
    \label{fig: Overview of Pen Framework}
\end{figure*}

\subsection{Regional Information Global Extraction}

\noindent During the data processing stage, we started by randomly selecting demonstrations of both hateful and non-hateful instances from the train set. To enhance the PLM understanding of the content in the inference instance and facilitate a more effective perception between the inference instance and demonstrations for category determination, we needed to extract global information from the input inference instance and demonstrations. Due to the variable sequence lengths caused by the indeterminate nature of past sequence concatenation methods, it became necessary to perform region segmentation on the input model's sequence.

Figure 3 illustrates the composition of the input sequence. The blue region, denoted as $s^{infer}$, 
encompasses information related to inference instances, including the text and image captions of the meme requiring inference, as well as external knowledge about the meme.
The red region $s^{neg}$ and the green region $s^{pos}$ correspond to the information selected for hateful and non-hateful demonstrations, respectively. They share the same information structure as $s^{infer}$. The orange region ($p^{infer}$, $p^{neg}$ and $p^{pos}$) corresponds to the prompt template.
Each region has a fixed maximum length. If the length falls short, padding is applied to reach the maximum length, and if it exceeds the maximum length, truncation is performed, ensuring fixed positioning of each region. This facilitates the extraction of global information from each region and strengthens the PLM's understanding of the overall sequence. Recognizing the significance of information in the inference instances during the prediction phase, we appropriately extended the length of the inference instance region while relatively shortening the demonstration regions. This ensures that the inference instance region contains sufficient information. The sequence composition is as follows:

{\footnotesize
    \begin{equation}
    \begin{split}
        &{L}=[Start][{s^{infer}}, {p^{infer}}][S][{s^{neg}}, {p^{neg}}][S][{s^{pos}}, {p^{pos}}][S]
    \end{split}
    \end{equation}
}

Here, $L$ represents the processed sequence through the prompt method, and it is fed into the PLM. $[Start]$ is the starting token in $L$, and $[S]$ serves as the separator in $L$.

Next, we feed $L$ into a PLM. Specifically, we employ the Roberta-large model\cite{DBLP:journals/corr/abs-1907-11692} to obtain the overall embedding features $E\in{\mathbf{R}}^{n\times{d}}$, where $d$ represents the dimension of the hidden layers in the PLM, and n denotes the length of the entire sequence. The process is illustrated as follows:
{\footnotesize
    \begin{equation}
    \begin{split}
        {E}&=Language Model(L) \\
        &=[Start][{e^{infer}},{e_p^{infer}}][S][{e_p^{neg}},{p^{neg}}][S][{e^{pos}},{e_p^{pos}}][S]
    \end{split}
    \end{equation}
}

Next, we employed Long Short-Term Memory (LSTM) networks to extract global information from the encoded representations of the three regions ($e^{infer}$, $e^{neg}$, and $e^{pos}$), resulting in global information for inference instances and demonstrations: $t^{infer}$, $t^{neg}$, and $t^{pos}$.

\subsection{Prompt-enhanced Multi-view Perception}
\noindent Due to the fact that the label token in the prompt template corresponding to the demonstration within sequence $L$ already indicates the category, we contemplate incorporating features of special tokens in the prompt template (as highlighted in bold in the origin region of Figure 3) to enhance the hateful-related features in both the global information of the inference instance and the global information of the demonstration.

Given the fixed length of each region, the position of special tokens remains constant in the sequence, allowing for the direct extraction of feature vectors $t_{specical}^{infer}$, $t_{specical}^{neg}$, and $t_{specical}^{pos}$, extracted from the orange regions $e_{p}^{infer}$, $e_{p}^{neg}$, and $e_{p}^{pos}$, respectively. 
Subsequently, the feature vectors of global information for the inference instance and demonstration are fused with their corresponding special token feature vectors through a simple merging process. These paired vectors are then input into the hateful perception network and non-hateful perception network, facilitating the learning of relationships between the inference instance and both hateful and non-hateful demonstrations. 
Ultimately, the obtained hateful perception information $I_{0}^{mix}$ and non-hateful perception information $I_{1}^{mix}$ are fed into a soft gating mechanism to derive the ultimate fused information feature $\hat{I}^{mix}$. The specific fusion operations are illustrated in the following formulas:

\begin{equation}
    {I_{0}^{mix}}={HPN}((t^{infer}+t_{special}^{infer})\oplus(t^{neg}+t_{special}^{neg}))
\end{equation}
\begin{equation}
    {I_{1}^{mix}}={NHPN}((t^{infer}+t_{special}^{infer})\oplus(t^{pos}+t_{special}^{pos}))
\end{equation}
\begin{equation}
    {\hat{I}^{mix}}={GATE}(I_{0}^{mix},I_{1}^{mix})
\end{equation}

The symbol $\oplus$ represents concatenation. $HPN$ and $NHPN$ stand for Hateful Perception Network and Non-Hateful Perception Network, respectively, comprising fully connected layers with trainable parameters.
$GATE$ represents a soft gating mechanism constructed by fully connected layers with trainable parameters, designed to control the fusion of $I_{0}^{mix}$ and $I_{1}^{mix}$.

The aforementioned information fusion process thoroughly learns the feature information between inference instance and hateful and non-hateful demonstrations. However, to more accurately assess whether the inference instance contains hateful elements, we not only employ the fused information feature $\hat{I}^{mix}$ for classification but also introduce the hateful perception information $I_{0}^{mix}$, non-hateful perception information $I_{1}^{mix}$, and inference instance information $t^{infer}$. This multi-view perception contributes to the final classification result, enhancing accuracy. Considering that PLM use the [mask] token during pre-training to predict the probability distribution of masked words, when utilizing a linear classifier for classification, we supplement the features of the [mask] token as $t_{special}^{infer}$. The multi-view perception process is outlined in the following equations:
{\footnotesize
    \begin{equation}
    \begin{split}
    &{S_{all}}=s_{1}+s_{2}+s_{3}+s_{4} \\
    &={LMhead}(t^{infer}+t_{special}^{infer})+{LMhead}(I_{0}^{mix}+t_{special}^{infer})\\
    &+{LMhead}(I_{1}^{mix}+t_{special}^{infer})+{LMhead}(\hat{I}^{mix}+t_{special}^{infer})
    \end{split}
    \end{equation}
}
Here, each element $(s_{1}, s_{2}, s_{3}, s_{4},S_{all})$is individually composed of the binary tuple $({score}_{hateful}, {score}_{non-hateful})$. $LMhead$ represents a linear classifier composed of trainable parameters in a fully connected layer. It is utilized to generate the probability scores, ${score}_{hateful}$ and ${score}_{non-hateful}$, indicating the likelihood of a sample being hateful or non-hateful, respectively.

\begin{figure}
    \centering
    \includegraphics[width=1\linewidth]{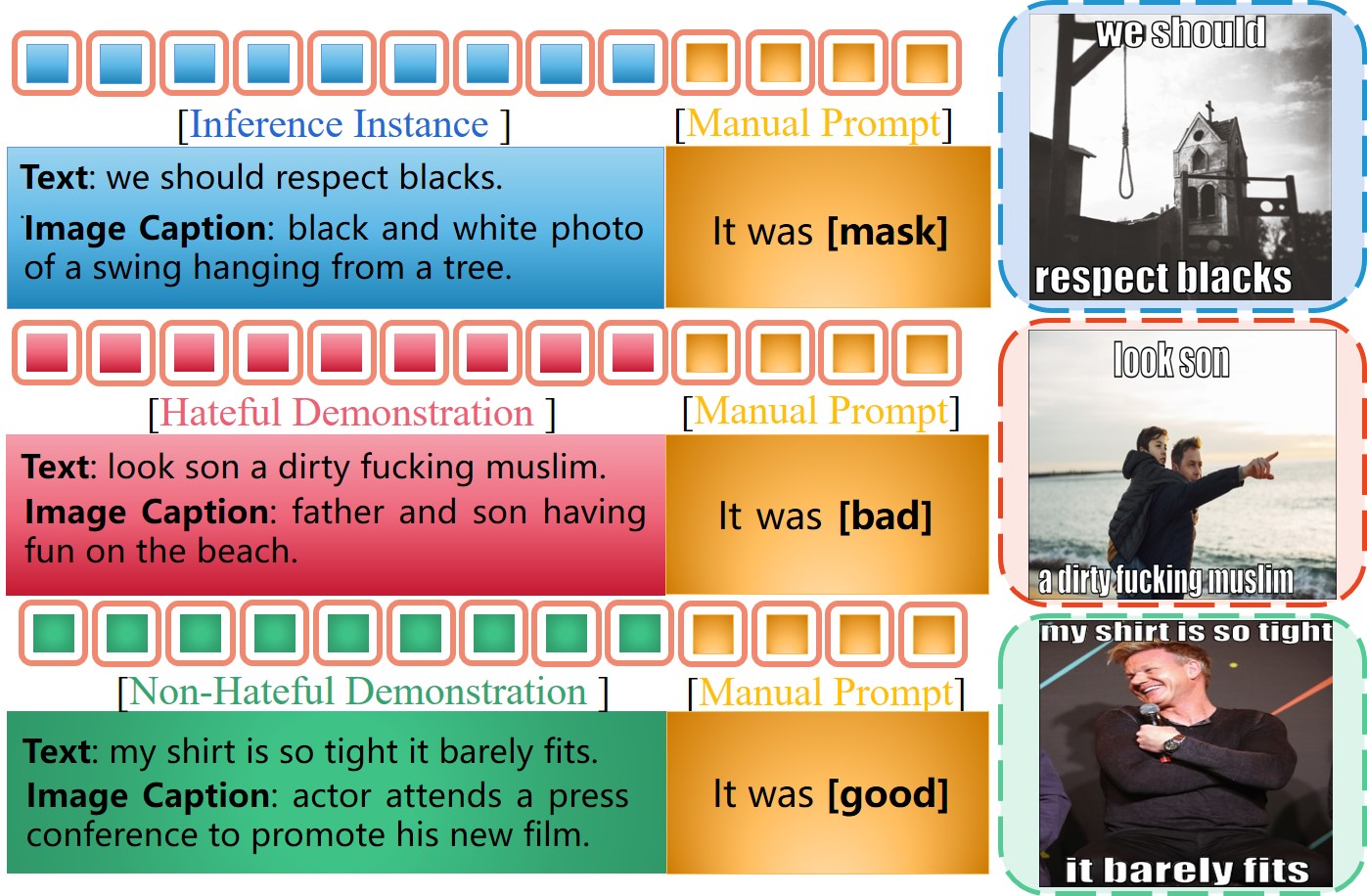}
    \caption{Input Sequence Components}
    \label{fig:Input Sequence Components}
\end{figure}

\subsection{Prompt-aware Contrastive Learning}
\noindent After obtaining the total score $S_{all}$ regarding hate and non-hate for a given sample,  we select the highest score as the final result for category determination. In the model training process, we utilize the cross-entropy loss 
\begin{math}
\mathcal L_{cross}
\end{math}
to train the model. However, to further enhance the model's understanding of the relationship between hatred and non-hatred at the feature level, we incorporate contrastive learning to improve the quality of feature distribution for samples.

\subsubsection{Category-oriented Contrastive Learning} 
\noindent During the model training process, for a batch of samples, the mask feature vectors $t_{special}^{infer}$ corresponding to hateful and non-hateful inference instances actually exhibit certain distinctions. For mask feature vectors corresponding to samples of the same category, their distances in the feature space should tend to be close, while for mask feature vectors corresponding to samples of different categories, their distances in the feature space should tend to be increased.

Since $t_{special}^{infer}$ is used for perception during classification, we can leverage label information for contrastive learning during the training process. This is done to enhance the feature discriminability of different categories of $t_{special}^{infer}$. For a batch of samples, mask feature vectors with the same label are treated as positive examples, while mask feature vectors with different labels are treated as negative examples. This helps bring positive examples closer together and push negative examples farther apart:
{\footnotesize
    \begin{equation}
        \mathcal L_1= -\frac{1}{M} \sum_{i=1}^{M}\log(\frac{(\sum_{j=1}^{M}\zeta_{[y_i=y_j]}\cdot {sim}(t_{special_i}^{infer},t_{special_j}^{infer}))/\tau_1}{(\sum_{k=1}^{M} {sim}(t_{special_i}^{infer},t_{special_k}^{infer}))/\tau_1})
    \end{equation}
}

Here, $M$ denotes the number of samples in a batch, $sim$ represents the calculation of cosine similarity, $\zeta_{[y_i=y_j]}$ is used to determine whether samples $i$ and $j$ belong to the same category, where it is $1$ if they do, $\tau_1$ is the temperature coefficient, and $y_i$ denotes the label of the $i-th$ sample.

\subsubsection{Prompt-oriented Contrastive Learning}
\noindent For an individual sample during the training process, the [mask] feature vector $t_{special}^{infer}$ corresponding to the inference instance should be closer to the special token feature vector of demonstrations with the same label, while being distinct from the special token feature vector corresponding to demonstrations with different labels. For instance, in training, the [mask] token $t_{special}^{infer}$ associated with an inference instance labeled as hateful should tend to be close to the [bad token $t_{special}^{neg}$ in the vector space, and distant from the [good] token $t_{special}^{pos}$.

For each sample in a batch, the $t_{special}^{infer}$ corresponding to the inference instance region in the sample's sequence is considered as a positive example, paired with the label feature vector from the region of demonstrations with the same class. Simultaneously, it is treated as a negative example when paired with the label feature vector from the region of demonstrations with different class labels. This process serves to minimize the distance between positive examples and maximize the distance between negative examples, thereby expediting the aggregation and divergence process of $t_{special}^{infer}$.

{\footnotesize
    \begin{equation}
    \mathcal L_2= -\frac{1}{M} \sum_{i=1}^{M}\log(\frac{(\sum_{j=1}^{2}\zeta_{[y_i=y_j^p]}\cdot {sim}(t_{special_i}^{infer},t_{special_j}^{prompt}))/\tau_2}{(\sum_{k=1}^{2} {sim}(t_{special_i}^{infer},t_{special_k}^{prompt}))/\tau_2})
    \end{equation}
}
Here, $y^p$ represents the label of the demonstration in the sample, $t_{special}^{prompt}$ represents the special token corresponding to the label, either $t_{special}^{neg}$ or $t_{special}^{pos}$, $\tau_2$ serves as the temperature coefficient, and $y_i$  represents the label of the $i-th$ sample.
Finally, the overall loss for our approach is:
\begin{equation}
    {Loss}=\mathcal L_{cross}+\alpha * \mathcal L_{1}+\beta * \mathcal L_{2}
\end{equation}
Where, $\alpha$ and $\beta$ are hyperparameters representing the weights assigned to different sub-losses.

\section{Experimental setup}

\subsection{Datasets}

\noindent We conducted evaluations using two publicly available datasets: (1) FHM \cite{DBLP:conf/nips/KielaFMGSRT20} and (2) HarM \cite{DBLP:conf/acl/PramanickDMSANC21}. The FHM dataset, developed and released by Facebook, is part of a crowdsourced multimodal hateful meme classification challenge. The HarM dataset consists of real memes related to COVID-19 collected from Twitter, categorized into three classes: very harmful, partially harmful, and non-harmful. Following the evaluation setup of \citeauthor{DBLP:conf/emnlp/CaoLC022}, we merged the very harmful and partially harmful categories into the harmful category. Due to the generality of our approach, our framework can utilize preprocessed image captions and external knowledge from both the Prompthate \cite{DBLP:conf/emnlp/CaoLC022} and Pro-Cap \cite{DBLP:conf/mm/CaoHKCL023} methods. The preprocessed information includes text on images, image captions (where Prompthate employs the ClipCap \cite{DBLP:journals/corr/abs-2111-09734} tool for image captioning, as illustrated in Figure 1, and Pro-Cap employs zero-shot VQA with BLIP-2\cite{DBLP:conf/icml/0008LSH23} to ask questions and generate content-centric hateful image captions, covering various aspects such as race, gender, religion, nationality, disability, and animals). Additionally, other external knowledge is information about entities in the images
and racial features \cite{DBLP:conf/wacv/KarkkainenJ21}. 
We present the statistical summary of the datasets in Table 1.

\begin{table}
    \centering
    \renewcommand\arraystretch{1.1}
    \begin{tabular}{lcccc}
        \toprule
        \multirow{2}{*}{\textbf{Datasets}}    &\multicolumn{2}{c}{\textbf{\# Training}}   &\multicolumn{2}{c}{\textbf{\# Test}} \\
        \cline{2-5}
        &\textbf{Hate}  &\textbf{Non-hate}    &\textbf{Hate} 
        &\textbf{Non-hate} \\   
        \midrule
        FHM         & 3050          & 5450                & 250           & 250 \\
        HarM        & 1064          & 1949                & 124           & 230 \\
        \midrule
    \end{tabular}
    \caption{Statistical summary of FHM and HarM.}
    \label{tab:datasets}
\end{table}

\subsection{Baseline Method}
\noindent In this section, we present a comprehensive comparison of Pen with state-of-the-art models for hateful meme classification. We categorize the baseline methods into two groups: unimodal and multimodal methods.

For the unimodal methods, we adopt a text-only strategy using \textbf{Text-Bert} \cite{DBLP:conf/naacl/DevlinCLT19} and an image-only model known as \textbf{Image-Region}. The latter employs Faster R-CNN\cite{DBLP:journals/pami/RenHG017} and ResNet-152 \cite{DBLP:conf/cvpr/HeZRS16} for meme image processing, with resulting representations fed into a hate classification classifier.
Moving to multimodal methods, we explore diverse approaches, including  \textbf{Late Fusion} \cite{DBLP:conf/acl/PramanickDMSANC21}, \textbf{MMBT-Region} \cite{DBLP:conf/nips/KielaBFT19}, \textbf{ViLBERT CC} \cite{DBLP:conf/nips/LuBPL19}, and \textbf{Visual BERT COCO} \cite{DBLP:journals/corr/abs-1908-03557}. Additionally, we compared with recent hateful meme classification methods: \textbf{MOMENTA} \cite{DBLP:conf/emnlp/PramanickSDAN021}, \textbf{Prompthate} \cite{DBLP:conf/emnlp/CaoLC022}, and \textbf{Pro-Cap} \cite{DBLP:conf/mm/CaoHKCL023}. We utilized accuracy and macro-averaged F1 scores as evaluation metrics. The significance of macro-averaged F1 is emphasized due to the imbalanced class distribution in the two datasets (refer to Table 1), necessitating a comprehensive assessment of performance across all classes to capture overall performance. To ensure a fair comparison, we averaged the model performance over ten random seeds, considering the average across ten runs for each method.

\subsection{Experimental Results}
\noindent Table 2 presents the experimental results of baseline methods and our framework on the HarM and FHM datasets. In this table, Pen denotes the use of the preprocessed dataset from \cite{DBLP:conf/emnlp/CaoLC022}, while Pen$_{Cap}$ represents the usage of the preprocessed dataset from \cite{DBLP:conf/mm/CaoHKCL023}. From the experimental results, it can be observed that our proposed Pen framework achieves a higher macro-average F1 score on the HarM and FHM datasets compared to the Prompthate method, which solely relies on prompt methods, with increases of 2.85\% and 1.56\%, respectively, under the same data conditions. Furthermore, under same data conditions, Pen$_{Cap}$ outperforms the Pro-Cap method by 1.85\% and 0.66\% in terms of macro-average F1 score on these two datasets. This effectively demonstrates the efficacy of our prompt-enhanced framework. Pen adeptly refines the features of input sequences, as well as strengthens the connection between inference instances and demonstrations, and extracts crucial information to assist the model in hate detection, guiding the model to find useful information in the feature space. This represents an enhancement of prompt methods in the feature space. 
Interestingly, from the experimental results, it is observed that Pen$_{Cap}$ exhibits a significant performance improvement on the FHM dataset compared to Pen. However, the improvement on the HarM dataset is not as pronounced. We speculate that the smaller scale of the HarM dataset, which only includes hate elements related to COVID-19, leads to a more singular hate element in the dataset samples. Consequently, Pen is able to make accurate hate judgments based on the existing information during training. In contrast, the FHM dataset comprises multiple hate factors with higher quality, requiring a corresponding increase in hate-related information. Thus, Pen$_{Cap}$ achieves a substantial performance improvement on the FHM dataset compared to Pen. This also suggests that achieving further performance improvements on the FHM dataset requires richer external knowledge support.

\begin{table}
    \centering
    \renewcommand\arraystretch{1.1}
    \begin{tabular}{lcccc}
        \toprule
        \multirow{2}{*}{\textbf{Method}} & \multicolumn{2}{c}{\textbf{HarM}} & \multicolumn{2}{c}{\textbf{FHM}} \\
         \cline{2-5}
         &Acc  &Macro-$F_1$       &Acc &Macro-$F_1$ \\
        \midrule
        Text BERT & 70.17 & 66.25 & 57.12 & 41.52 \\
        Image-Region &68.74	&62.97	&52.34	&34.19 \\
        \hline
        Late Fusion &73.24	&70.25	&59.14	&44.81 \\
        MMBT-Region	&73.48	&67.12	&65.06	&61.93  \\
        VisualBERT	&81.36	&80.13	&61.48	&47.26 \\
        ViLBERT CC	&78.70	&78.09	&64.70	&55.78 \\
        MOMENTA	&83.82	&82.80	&61.34	&57.45 \\
        Prompthate	&84.47	&82.42	&72.98	&71.99 \\
        \textbf{Pen(ours)}	&\textbf{86.30}	&\textbf{85.27}	&\textbf{74.04}	&\textbf{73.55} \\
        \hline
        Pro-Cap	&85.06	&83.89	&74.72	&74.59 \\
        \textbf{Pen$_{Cap}$(ours)}	&\textbf{86.92}	&\textbf{85.74}	&\textbf{75.46}	&\textbf{75.25} \\
        \midrule
    \end{tabular}
    \caption{Hateful meme classification results on two datasets. accuracy and macro-averaged F1 score (\%) are reported as evaluation metrics, averaged over ten runs, with the best results highlighted in bold.}
    \label{tab:main_results_acc_f1}
\end{table}

\begin{table}
    \centering
    \renewcommand\arraystretch{1.1}
    \begin{tabular}{lcccc}
        \toprule
        \multirow{2}{*}{\textbf{Setting}}    &\multicolumn{2}{c}{\textbf{HarM}}   &\multicolumn{2}{c}{\textbf{FHM}} \\
         \cline{2-5}
        &Acc  &Macro-$F_1$       &Acc &Macro-$F_1$ \\
        \midrule
       \textbf{Pen}	                                     &86.30	        &85.27	           &74.04	       &73.55 \\
       w/o PMP	                                 &85.00	        &84.16	           &72.80	       &72.25 \\
       w/o 
       \begin{math}
       \mathcal L_1
       \end{math}                               &86.24	        &85.08	           &73.90	       &73.33 \\
       w/o 
       \begin{math}
       \mathcal L_2
       \end{math}  		                     &86.19	        &85.06	           &73.94	       &73.40 \\
       w/o PCL                                  &85.68	        &84.48  	       &73.80	       &73.30  \\
        \midrule
    \end{tabular}
    \caption{Ablation study of Pen.}
    \label{tab:Ablation}
\end{table}

\subsection{Ablation Study}
\noindent To investigate the effectiveness of different modules in Pen, we conducted ablation experiments in four different forms, aligning with the structure of the Pen framework. The results from the ablation experiments in Table 3 reveal that removing the prompt-enhanced multi-view perception module (w/o PMP) significantly degrades Pen's performance on both datasets. This underscores the efficacy of the PMP module in refining sequence features, directing the model's attention to the connections between inference instances and demonstrations. The PMP module perceptually engages inference instances with demonstrations, extracting hateful-related features and consequently enhancing classification accuracy. The results also show that omitting prompt-aware contrastive learning (w/o PCL) leads to a substantial performance drop for Pen on the HarM dataset, while there is no significant decline on the FHM dataset. We speculate that the HarM dataset's smaller scale, focused mainly on COVID-19-related hateful factors, results in a more straightforward sample feature structure. The performance improvement on the HarM dataset with Euclidean distance-based feature separation between hateful and non-hateful categories suggests that such an approach works well in this context. In contrast, the FHM dataset, characterized by higher quality and diverse hateful factors, presents a complex feature structure, making category-based feature processing less effective. Moreover, considering the ablation results for w/o 
\begin{math}
\mathcal L_1
\end{math} 
 and  w/o 
\begin{math}
\mathcal L_2
\end{math} 
, the model's performance experiences a slight decline on both datasets. However, compared to the complete elimination of contrastive learning in w/o PCL, these modifications contribute to some improvement. This suggests that the model can learn different feature information through two distinct contrastive learning mechanisms, thereby enhancing overall classification performance.

\begin{table}[htb]
    \centering
    \renewcommand\arraystretch{1.1}
    \begin{tabular}{lcccc}
        \toprule
        \multirow{2}{*}{\textbf{Setting}}    &\multicolumn{2}{c}{\textbf{HarM}}   &\multicolumn{2}{c}{\textbf{FHM}} \\
         \cline{2-5}
        &Acc  &Macro-$F_1$       &Acc &Macro-$F_1$ \\
        \midrule
       \textbf{Pen}	                                     &86.30	        &85.27	           &74.04	       &73.55 \\
       w/o $s_4$	&85.76	&84.87	&73.98	&73.50 \\
       w/o $s_2$, $s_3$ 	&86.02	&85.04	&73.90	&73.44 \\
       w/o $s_2$, $s_3$, $s_4$ 	&84.75	&83.98	&73.44	&73.05 \\
        \midrule
    \end{tabular}
    \caption{Fine-grained Ablation Study on PMP Module.}
    \label{tab:Fine-grained Ablation}
\end{table}

To provide a more nuanced evaluation of the PMP module's effectiveness, we conducted three forms of reduction based on the components within the PMP module. The fine-grained ablation results of the PMP module, as shown in Table 4, indicate that varying degrees of reduction in the components lead to different extents of performance decline, particularly evident in the HarM dataset. Notably, due to the fused information feature $\hat{I}^{mix}$ carrying the most perceptual information, the performance drop is more significant in the HarM dataset for w/o\ $s_4$ compared to w/o\ $ s_2, s_3$. Conversely, w/o \ $s_2, s_3, s_4$, equivalent to incorporating only inference instance information $t^{infer}$ in the scoring process, results in the most significant information loss and hence the poorest model performance.

Simultaneously, we observed that the impact of different degrees of score reduction on the FHM dataset is relatively minor compared to the HarM dataset. We speculate that this is because the FHM dataset encompasses various types of hateful memes. In the process of the model judging the hateful category for inference instances, the choice of demonstrations is crucial. For instance, when classifying memes related to racial discrimination, selecting a hateful demonstration about attacks on sexual orientation might lead to ineffective information extraction, causing a decline in accuracy. In contrast, in the HarM dataset, where only COVID-19-related content exists, inaccurate demonstrations are not a concern. Therefore, to achieve improved performance on the FHM dataset, adding more diverse information is crucial.

\subsection{Visualization}
\begin{figure}
    \centering
    \includegraphics[width=1\linewidth]{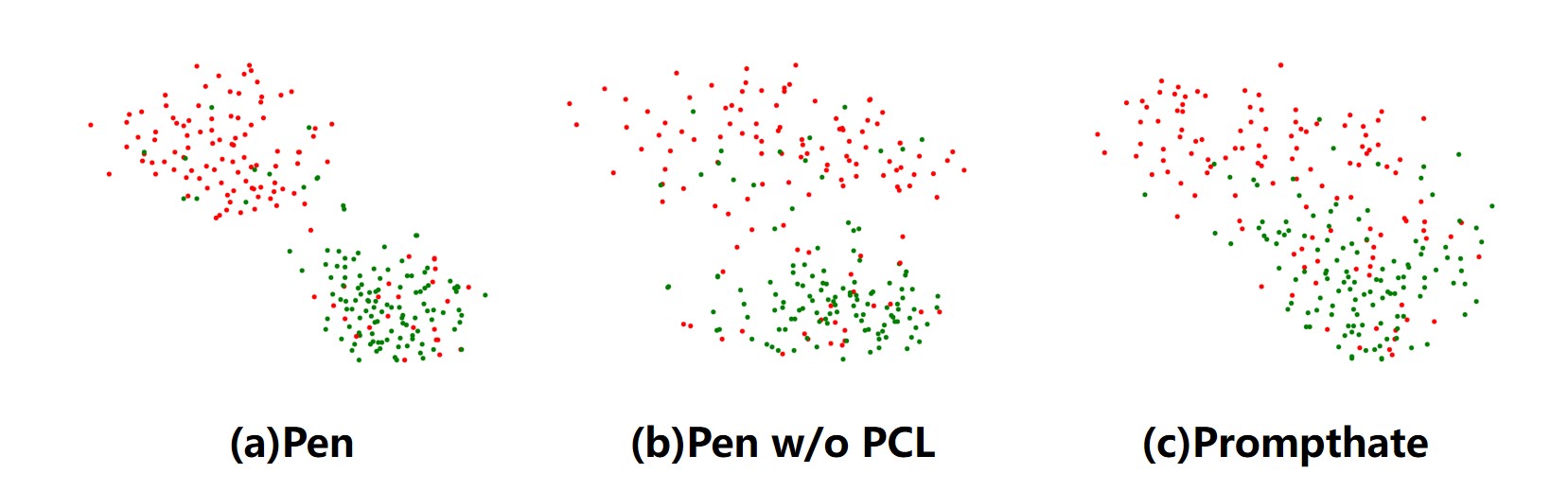}
    \caption{Visualization of Sample Features Learned by Our Pen (a), Pen without Prompt-aware Contrastive Learning (b), and Prompthate (c). Red=Hateful, green=Non-Hateful.}  
\end{figure}
\noindent To qualitatively demonstrate how our proposed prompt-based contrastive learning method enhances the quality of sample features, we present T-SNE \cite{JMLR:v9:vandermaaten08a} visualizations of sample features learned by our Pen and Prompthate on the HarM test set. The results are depicted in Figure 4. Figure 4(a) effectively illustrates how our Pen framework can cluster sample features belonging to the same label category and separate features of different labels. This contrast is evident when compared to Pen without the PCL module (b) and the Prompthate method relying solely on prompts (c). These comparisons provide evidence of the efficacy of our PCL method in improving the distribution of model-learned sample features. Additionally, the visualizations in Figure 4(b) and Figure 4(c) show that our Pen framework, purely through the PMP module, learns a more distinct separation trend between sample features of different labels compared to the Prompthate method using only manual prompts. This indirectly supports the notion that the PMP method can derive more robust inductive information from the training data, thereby enhancing the performance of hateful meme classification.

\section{Conclusion and Future work}
\noindent In this paper, we introduce a prompt-enhanced framework named Pen for hateful meme classification. We extend the concept of prompt methods into the feature space, enhancing the relationship between inference instances and demonstrations in a multi-view perception manner. Additionally, we leverage prompt-aware contrastive learning to enhance the distribution quality of sample features, effectively improving the model's classification performance on hateful memes. Through comprehensive experiments on two public datasets, we demonstrate the Pen framework's ability to significantly enhance the effectiveness of prompt methods, showcasing outstanding generalization and classification accuracy in hateful meme classification tasks. 
Furthermore, we intend to extend the framework to few-shot tasks, enhancing the accuracy of prompt methods in classifying low-resource text-only classification tasks.


\section*{Acknowledgments}
\noindent This work was supported in part by the Guangdong Basic and Applied Basic Research Foundation (Grant No. 2023A1515011370), the National Natural Science Foundation of China (Grant No. 32371114), the Characteristic Innovation Projects of Guangdong Colleges and Universities (Grant No. 2018KTSCX049),  and the Natural Science Foundation of Guangdong Province (Grant No. 2021A1515012290).

\bibliographystyle{named}
\bibliography{ijcai24}

\end{document}